\documentclass[conference]{IEEEtran}
\IEEEoverridecommandlockouts
\usepackage{amsmath,amssymb,amsfonts}
\usepackage{algorithmic}
\usepackage{graphicx}
\usepackage{booktabs} 
\usepackage{textcomp}
\usepackage{xcolor}
\usepackage{comment}
\usepackage[backend=biber,style=ieee,url=true,natbib=true]{biblatex}
\usepackage[hidelinks]{hyperref}
\def\BibTeX{{\rm B\kern-.05em{\sc i\kern-.025em b}\kern-.08em
    T\kern-.1667em\lower.7ex\hbox{E}\kern-.125emX}}
\begin{document}

\title{Assessment of Evolving Large Language Models in Upper Secondary Mathematics\\
\thanks{Support from the Research Council of Finland under Grant number 353325 is gratefully acknowledged.}
}

\author{\IEEEauthorblockN{1\textsuperscript{st} Mika Setälä}
\IEEEauthorblockA{\textit{Faculty of Information Technology} \\
\textit{University of Jyväskylä}\\
Jyväskylä, Finland \\
0009-0006-7195-9587}
~\\
\and
\IEEEauthorblockN{2\textsuperscript{nd} Pieta Sikström*}
\IEEEauthorblockA{\textit{Faculty of Information Technology} \\
\textit{University of Jyväskylä}\\
Jyväskylä, Finland \\
0000-0002-2055-7995}
*Corresponding author
~\\
\and
\IEEEauthorblockN{3\textsuperscript{rd} Ville Heilala}
\IEEEauthorblockA{\textit{Faculty of Humanities and Social Sciences} \\
\textit{University of Jyväskylä}\\
Jyväskylä, Finland \\
0000-0003-2068-2777}
~\\
\and
\IEEEauthorblockN{4\textsuperscript{th} Tommi Kärkkäinen}
\IEEEauthorblockA{\textit{Faculty of Information Technology} \\
\textit{University of Jyväskylä}\\
Jyväskylä, Finland \\
0000-0001-5469-5379}
}

\maketitle

\begin{abstract}
Large language models (LLMs) have shown increasing promise in educational settings, yet their mathematical reasoning has been considered evolving. This study evaluates the mathematical capabilities of various LLMs using the Finnish matriculation examination, a high-stakes digital test for upper secondary education. Initial tests yielded moderate performance corresponding to mid-range grades, but later evaluations demonstrated substantial improvements as the language models evolved. Remarkably, some models achieved near-perfect or perfect scores, matching top student performance and qualifying for university admission. Our findings highlight the rapid advances in the mathematical proficiency of LLMs and illustrate their potential as underlying tools to support learning and teaching in a variety of ways.
\end{abstract}


\section{Introduction}

Conversational learning technologies have demonstrated significant potential in educational contexts across a wide range of disciplines \citep{sikstrom2022pedagogical,sikstrom2024pedagogical,Heilala2025-wg}. 
Since OpenAI launched ChatGPT in late 2022, the education landscape has undergone disruption, where Large Language Model (LLM) based learning tools have been increasingly embedded in education. LLMs can enhance personalized learning \citep{pesovski2024generative} and provide individual scaffolding for students while enhancing engagement through interactive learning experiences \citep{kasneci2023chatgpt}. The findings from a literature review \citep{almarashdi2024unveiling} suggested that, in general, ChatGPT can enhance personalized learning, motivation, and engagement. However, also critical views have been expressed on whether the summarization capabilities of generative AI are used to outsource learning to a machine instead of scaffolding student's own learning \citep{corbin2025missing}.

The potential benefits of LLMs are also present in mathematics, where several studies have concluded the benefits of using conversational AI \citep{frieder2023mathematical,dasari2024chatgpt,urhan2024argumentation,zhao2023survey,wardat2023chatgpt}. For example, a conversational agent can both answer mathematical questions and also serve as a mathematical search engine and knowledge base \citep{frieder2023mathematical}. With the growing recognition of the benefits and increasing prevalence of LLMs, a deeper understanding of their role in education is essential  \citep{siegle2023twenty}.  In particular, given their somewhat ambiguous capabilities in mathematical reasoning \citep{ahn2024large}, it is crucial to explore their impact and effectiveness in mathematics learning.

Digital performance in Finland is distinguished\footnote{\href{https://digital-strategy.ec.europa.eu/en/library/digital-economy-and-society-index-desi-2022}{Digital Economy and Society Index 2022 - Country Reporting}}, and nearly 80\% of the learning materials in upper secondary education were digital in 2023 \citep{OECDDigitalization2023}. The national matriculation examination\footnote{\href{https://www.ylioppilastutkinto.fi/en/matriculation-examination}{The Finnish Matriculation Examination}} taken at the end of the Finnish upper secondary school (ISCED Level 3) is a fully digitalized, biannual, high-stakes test designed to assess how well students have met curriculum requirements \citep{virtanen2024vector}. Since rigorous benchmarking of LLMs relies on standardized evaluation criteria to ensure fair and objective comparison between models, the aim of this study was to contribute to the understanding of LLMs' mathematical reasoning capabilities by using the Finnish matriculation examination in advanced mathematics as test data. To explore the mathematical performance of LLMs, we pose the following research question: \textit{How do different LLM models perform in Finnish matriculation examination?}

\section{Capabilities of LLMs in mathematics }

\begin{table*}[htbp!]
    \centering
    \caption{Versions and initial release dates of OpenAI\textsuperscript{a}, Google\textsuperscript{b}, and DeepSeek\textsuperscript{c} models}
    \begin{tabular}{p{2.8cm}p{1.5cm}p{2cm}p{6.5cm}}
        \toprule
        \textbf{LLM Model} & \textbf{Initial Release Date} & \textbf{Initial Knowledge Cutoff} & \textbf{Description} \\
        \midrule
        ChatGPT (GPT-3.5)\textsuperscript{a} & \href{https://web.archive.org/web/20221130180912/https://openai.com/blog/chatgpt/}{2022/11/30} & \href{https://web.archive.org/web/20231112181303/https://help.openai.com/en/articles/8555514-gpt-3-5-turbo-updates}{2021/09} & The initial OpenAI ChatGPT release, a sibling model to \href{https://web.archive.org/web/20220130195139/https://github.com/openai/following-instructions-human-feedback/blob/main/model-card.md}{InstructGPT} \\
        \href{https://cdn.openai.com/papers/gpt-4-system-card.pdf}{GPT-4}\textsuperscript{a} & \href{https://web.archive.org/web/20230314165432/https://openai.com/research/gpt-4}{2023/03/14} & \href{https://web.archive.org/web/20230314165432/https://openai.com/research/gpt-4}{2021/09} & A large multimodal model introduced with the ChatGPT Plus subscription \\
        \href{https://web.archive.org/web/20230321140901/https://ai.google/static/documents/google-about-bard.pdf}{Bard\textsuperscript{b}} & \href{https://web.archive.org/web/20230321140848/https://blog.google/technology/ai/try-bard/}{2023/03/21} & & Google's chatbot initially based on the LaMDA language model. \\
        \href{https://doi.org/10.48550/arXiv.2401.14196}{DeepSeek Coder\textsuperscript{c}} & \href{https://web.archive.org/web/20250213083110/https://huggingface.co/deepseek-ai/deepseek-coder-33b-instruct/commit/4e3839a083fabecedcdeb1481c79baea685bb4c8}{2023/11/01} & & Initial release focusing on coding capabilities \\
        GPT-4 Turbo\textsuperscript{a} & \href{https://web.archive.org/web/20231106215514/https://openai.com/blog/new-models-and-developer-products-announced-at-devday}{2023/11/06} & \href{https://web.archive.org/web/20231106215514/https://openai.com/blog/new-models-and-developer-products-announced-at-devday}{2023/04} & A model with a 128k context window \\
        \href{https://doi.org/10.48550/arXiv.2401.02954}{DeepSeek LLM\textsuperscript{c}} & \href{https://web.archive.org/web/20240627225000/https://huggingface.co/deepseek-ai/deepseek-llm-67b-chat/tree/main}{2023/11/29} & & Expanded language model with improved performance \\
        \href{https://web.archive.org/web/20231206151650/https://storage.googleapis.com/deepmind-media/gemini/gemini_1_report.pdf}{Gemini 1\textsuperscript{b}} & \href{https://web.archive.org/web/20231206170410/https://blog.google/technology/ai/gemini-collection/}{2023/12/06} & & Bard rebranded to Gemini and updated to the Gemini LLM\\
        \href{https://doi.org/10.48550/arXiv.2401.06066}{DeepSeek-MoE\textsuperscript{c}} & \href{https://web.archive.org/web/20240111233223/https://huggingface.co/deepseek-ai/deepseek-moe-16b-base/tree/main}{2024/01/09} & & Mixture-of-Experts architecture for more efficient modeling \\
        \href{https://doi.org/10.48550/arXiv.2403.05530}{Gemini 1.5\textsuperscript{b}} & \href{https://web.archive.org/web/20240215151350/https://blog.google/technology/ai/google-gemini-next-generation-model-february-2024/}{2024/02/15} & & A new architecture and expanded context handling\\
        \href{https://doi.org/10.48550/arXiv.2405.04434}{DeepSeek V2\textsuperscript{c}} & \href{https://web.archive.org/web/20240623020441/https://huggingface.co/deepseek-ai/DeepSeek-V2/tree/main}{2024/05/06} & & Introduced new features and enhancements \\
        \href{https://cdn.openai.com/gpt-4o-system-card.pdf}{GPT-4o\textsuperscript{a}} & \href{https://web.archive.org/web/20240513173656/https://openai.com/index/hello-gpt-4o/}{2024/05/13} & \href{https://web.archive.org/web/20240516092808/https://learn.microsoft.com/en-us/azure/ai-services/openai/concepts/models}{2023/10} & Multimodal model capable of processing text, images, and audio \\ 
        \href{https://web.archive.org/web/20240912185132/https://assets.ctfassets.net/kftzwdyauwt9/2pON5XTkyX3o1NJmq4XwOz/a863fd35000b514887366623a5738b83/o1_system_card.pdf}{o1-preview\textsuperscript{a}} & \href{https://web.archive.org/web/20240912171617/https://openai.com/index/introducing-openai-o1-preview/}{2024/09/12} & \href{https://web.archive.org/web/20241003110359/https://learn.microsoft.com/en-us/azure/ai-services/openai/concepts/models}{2023/10} & Chain-of-thought reasoning model \\ 
        \href{https://web.archive.org/web/20240912185132/https://assets.ctfassets.net/kftzwdyauwt9/2pON5XTkyX3o1NJmq4XwOz/a863fd35000b514887366623a5738b83/o1_system_card.pdf}{o1-mini\textsuperscript{a}} & \href{https://web.archive.org/web/20240912183657/https://openai.com/index/openai-o1-mini-advancing-cost-efficient-reasoning/}{2024/09/12} & \href{https://web.archive.org/web/20241003110359/https://learn.microsoft.com/en-us/azure/ai-services/openai/concepts/models}{2023/10} & Tunable reasoning effort for STEM fields, esp. in math and coding \\
        \href{https://web.archive.org/web/20241205184050/https://cdn.openai.com/o1-system-card-20241205.pdf}{o1\textsuperscript{a}} & \href{https://web.archive.org/web/20241205190605/https://openai.com/index/introducing-chatgpt-pro/}{2024/12/05} & \href{https://web.archive.org/web/20250114184854/https://learn.microsoft.com/en-us/azure/ai-services/openai/concepts/models?tabs=global-standard%2Cstandard-chat-completions}{2023/10} & A state-of-the-art reasoning model for problem-solving in various domains \\
        Gemini 2.0 Flash Exp\textsuperscript{b} & \href{https://web.archive.org/web/20241211154129/https://blog.google/technology/google-deepmind/google-gemini-ai-update-december-2024/}{2024/12/11} & & Enhancements in speed and performance \\
        \href{https://doi.org/10.48550/arXiv.2412.19437}{DeepSeek V3\textsuperscript{c}} & \href{https://web.archive.org/web/20241225153515/https://huggingface.co/deepseek-ai/DeepSeek-V3-Base/tree/main}{2024/12/25} & & Performance improvements with multimodal support\\
        \href{https://doi.org/10.48550/arXiv.2501.12948}{DeepSeek R1\textsuperscript{c}} & \href{https://web.archive.org/web/20250120073339/https://huggingface.co/deepseek-ai/DeepSeek-R1/tree/main}{2025/01/20} & & Logical inference and reasoning for mathematical problem-solving \\
        Gemini 2.0 Flash\textsuperscript{b} & \href{https://web.archive.org/web/20250130214636/https://blog.google/feed/gemini-app-model-update-january-2025/}{2025/01/30} & & Adjusted reasoning with large context window \\
        \href{https://web.archive.org/web/20250131182131/https://cdn.openai.com/o3-mini-system-card.pdf}{o3-mini\textsuperscript{a}} & \href{https://web.archive.org/web/20250131194506/https://openai.com/index/openai-o3-mini/}{2025/01/31} & \href{https://web.archive.org/web/20250207014702/https://learn.microsoft.com/en-us/azure/ai-services/openai/concepts/models?tabs=global-standard%2Cstandard-chat-completions}{2023/10} & A reasoning model particularly for STEM subjects in science, math, and coding \\

        \bottomrule
    \end{tabular}
    \label{tab:versions}
\end{table*}

The rapid adoption of LLMs has fueled competition among providers, leading to an unprecedented acceleration in development. In 2023 alone, 21 major LLMs were released \citep{naveed2023comprehensive}, and the speed has been increasing ever since. Table \ref{tab:versions} presents an overview of the development trajectory and different versions of some popular LLMs.

Experiences and experiments on the use of LLMs in mathematics education are conditioned on the mathematical capabilities of these systems. For example, ChatGPT's mathematical competence is argued to be sufficient to teach rational numbers according to middle school curricula \citep{kaplan2025chatgpt}. However, it lacked pedagogical content knowledge, particularly in explaining the representation of rational numbers on a number line and their conversion to decimal form. Some argued that the use of ChatGPT in mathematical reasoning requires expertise in calculus concepts \citep{urhan2024argumentation}. In addition, in engineering mathematics, certain challenges necessitated the adaptation of teaching strategies and methodologies \citep{sanchez2023chatgpt}. 

\citet{frieder2023mathematical} reported that the exam-solving capabilities of \textsc{ChatGPT/GPT-4} were well below the level of a graduate student. In addition, according to tests by \citet{achiam2023gpt} in January 2023, \textsc{ChatGPT} was not able to produce high-quality proofs or calculations consistently, even if the quality of the answers was sometimes surprisingly good. These observations were reflected by \citet{zhao2023survey}, who concluded that mathematical reasoning and coding abilities should be enhanced by training with more mathematical texts and code data. In October 2024, Apple researchers \citep{mirzadeh2024gsm} presented findings that LLMs fail to perform genuine logical reasoning in mathematical tasks. Their study revealed that even minor modifications in the formulation of a problem could significantly degrade model performance.

Previous assessments of AI mathematical proficiency have predominantly relied on three key datasets (Table \ref{tab:math_benchmark}). MATH Dataset \citep{hendrycks2021math} is a collection of 12,500 problems sourced from high school and undergraduate mathematics competitions. The subset MATH-500 comprises the 500 most challenging problems, widely used to evaluate LLM performance in mathematical reasoning. American Invitational Mathematics Examination (AIME) \citep{maa_invitationals} is a 15-question, three-hour test that has been administered to students scoring in the top 5\% of the AMC 12 competition. AIME serves as an intermediate stage in the selection process for the USAMO and the USAJMO, which are key qualifiers for the International Mathematical Olympiad. The difficulty of AIME problems increases progressively, and each answer is a three-digit integer. The problems span various mathematical fields, including algebra, geometry, number theory, and combinatorics, demanding advanced problem-solving skills and creativity. Chinese National Mathematics Olympiad (CNMO) serves as a selection process for the national team competing in the International Mathematical Olympiad. It focuses on deep mathematical concepts like functional equations, symmetries, and mathematical analysis. These benchmarks have provided a foundation for evaluating AI's mathematical abilities, yet new evaluation frameworks will be necessary to assess progress meaningfully as AI capabilities evolve. 

\begin{table}[htbp!]
    \centering
    \caption{Mathematical benchmark results \citep{guo2025,openai2025}}
    \begin{tabular}{p{1.5cm}p{0.5cm}p{0.5cm}p{0.5cm}p{0.5cm}p{0.5cm}p{0.5cm}}
        \toprule
        \textbf{Benchmark (Pass@1)} & \textbf{o3-mini (high)}& \textbf{4o 0513} & \textbf{V3} & \textbf{o1-mini} & \textbf{o1-1217} & \textbf{R1} \\
        \midrule
        AIME 2024 & 87.3& 9.3 & 39.2 & 63.6 & 79.2 & \textbf{79.8} \\
        MATH-500 & -& 74.6 & 72.3 & 91.4 & \textbf{97.3} & 96.3 \\
        CNMO 2024 & -& 10.8 & 43.2 & 67.6 & - & \textbf{78.8} \\
        \bottomrule
    \end{tabular}
    \label{tab:math_benchmark}
\end{table}

\section{Capabilities of LLMs in Finnish matriculation examination of mathematics}\label{subsec:GenAIPerf}

In this study, the mathematical capabilities of different LLM versions (See Table \ref{tab:versions}) were evaluated in four time periods: August 2023, November 2023, April 2024, and January 2025. The Finnish matriculation examination of mathematics is arranged at two different levels of difficulty: the advanced level and the basic level. This study used four versions of the \href{https://www.ylioppilastutkinto.fi/en/matriculation-examination/tests-examination}{Finnish matriculation examination in advanced mathematics level},  \href{https://info.ylioppilastutkinto.fi/hvp/final/2021_k_m.pdf}{Spring 2001}, \href{https://tiedostot.ylioppilastutkinto.fi/kokeet/2023-09-19_M_fi/grading-instructions.html}{Autumn 2023}, \href{https://tiedostot.ylioppilastutkinto.fi/kokeet/2024-03-20_M_fi/grading-instructions.html}{Spring 2024} and \href{https://tiedostot.ylioppilastutkinto.fi/kokeet/2024-09-26_M_fi/grading-instructions.html}{Autumn 2024}. The assessment was organized similarly to tests with GPT-4, which were based on examinations originally designed for humans \citep{achiam2023gpt}. The LLMs' answers were graded according to the national grading guidelines \citep{YoKoeGrading}. The Finnish matriculation examination is graded with seven grades: L, E, M, C, B, A, and I, where L refers to the highest grade and I refers to failing the examination \citep{YoKoeGrading}. 

In August 2023, the mathematical performance was tested by using the examination from Spring 2021. By this time, OpenAI's \textsc{GPT-4} scored 64 out of 120 points, achieving the second-highest grade of E. In contrast, Google's Bard scored 34 points, corresponding to a middle-grade C.

In November 2023, OpenAI's \textsc{GPT-4} was tested using the examination from Autumn 2023, and its performance was significantly improved because it utilized programming capabilities in Python. Symbolic calculation proceeded with the assistance of Python's \textsc{SymPy} library, and  OpenAI's \textsc{GPT-4} was able to visualize the complexities of numerous challenging space geometry tasks accurately. The score was 93 out of 120 points, corresponding to the highest grade L.

In April 2024, three different LLMs were evaluated using the spring 2024 examination. A summary of the free and paid versions of the three service providers: OpenAI \textsc{ChatGPT}, Microsoft's \textsc{Copilot}, and Google's \textsc{Gemini} is presented in Table \ref{tab:summary_scores_grades}. ChatGPT 4-turbo (paid) achieved L with 92/120 points, \textsc{Bard} (\textsc{Gemini}) (free version) scored 40/120 with a grade of C, and \textsc{Copilot} (paid) was at a level scoring 66/120 and E.

\begin{table}[htbp!]
	\centering
	\caption{Summary of total scores and grades for various models in the various MAA examinations (maximum score 120)}
	\label{tab:summary_scores_grades}
	\begin{tabular}{lccc}
		\toprule
		\textbf{Model (testing time)} &  \textbf{Exam}&\textbf{Score=Mem}& \textbf{Grade} \\ \midrule
            OpenAI GPT-4  (23/08/05) & S2021& 64&E\\  
            Google Bard (23/08/05) & S2021 & 34&C\\ 
            OpenAI GPT-4 (23/10/26) & A2023 & 93&L\\ 
            OpenAI gpt-4-turbo (24/04/09)& S2024 & 92& L\\  
		CoPilot PRO (24/04/30) & S2024&66  & E\\
		Google Gemini Free (24/04/30) & S2024 & 40  & C \\
 Google Gemini 2.0 (25/01/25)& A2024& 115&L\\
 ChatGPT o1 (25/01/25)& A2024& 118&L\\
 ChatGPT o3 (25/02/05)& A2024& 120&L\\
 DeepSeek R1 (25/01/27)& A2024& 120&L\\
		\bottomrule\\
	\end{tabular}
	Exam (Finnish matriculation examination): 
    S=Spring, A=Autumn, \\ 
    Grade: L (the best), E, M, C, B, A, I (fail) \\ 
\end{table}

In January 2025, the fourth evaluation was conducted using the autumn 2024 matriculation examination. At the time, OpenAI’s o1 model had already established itself as the leading contender. Still, the recent arrival of DeepSeek R1 outperformed o1 by achieving a perfect score across all tasks. However, OpenAI quickly responded with the release of o3, which matched DeepSeek’s performance. Meanwhile, the Gemini 2.0 Experimental Advanced version also significantly improved its capabilities, reaching a much higher performance level than in the previous evaluations of the model. In January 2025, results showed Gemini scoring 115/120, ChatGPT o1 reaching 118/120, and both DeepSeek R1 and ChatGPT o3 achieving the full 120/120.

\begin{figure}
    \centering
    \caption{Development of capabilities of two leading LLMs}
    \includegraphics[width=1\linewidth]{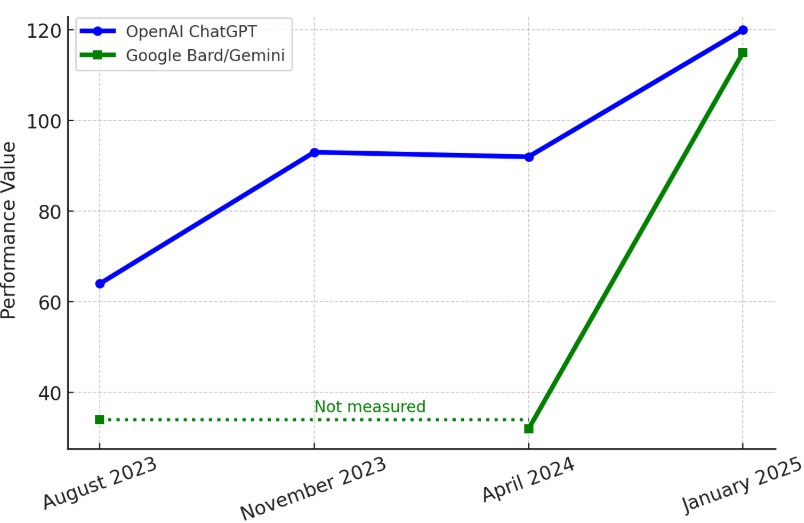}
    \label{lab:PerfComp}
\end{figure}

These exceptionally high scores indicated that the models had reached or even exceeded the limits of the current grading scale. As a result, the January evaluation was expanded to include all 13 tasks, enabling a more detailed analysis of each model’s full problem-solving capacity and relative strengths (Table~\ref{tab:comparison}). To provide a visual overview of the rapid progress observed, Figure \ref{lab:PerfComp} illustrates the development of mathematical performance over time for two leading LLMs. The figure highlights how the capabilities of these models evolved significantly from moderate results in August 2023 to perfect scores in January 2025.

\begin{table*}[htbp!]
    \centering
    \caption{Comparison of LLMs on Autumn 2024 MAA exam.}
    \begin{tabular}{lp{0.2cm}p{0.2cm}p{0.2cm}p{0.2cm}p{0.2cm}p{0.2cm}p{0.2cm}p{0.2cm}p{0.2cm}p{0.3cm}p{0.3cm}p{0.3cm}p{0.5cm}r}
        \toprule
        Model & T1 & T2 & T3 & T4 & T5 & T6 & T7 & T8 & T9 & T10 & T1 & T12 & T13 & Sum (G) \\ 
        \midrule
        ChatGPT o1& 11 & 12 & 12 & 12 & 12 & 12 & 12 & 12 & 12 & 6 & 6 & 12 & 10 & 141 (L) \\ 
        Gemini 2.0 exp. adv.& 12 & 12 & 8 & 12 & 10 & 12 & 12 & 9 & 12 & 4 & 8 & 12 & 12 & 135 (L) \\
        Deepseek R1& 12 & 12 & 12 & 12 & 12 & 12 & 12 & 12 & 12 & 12 & 12 & 12 & 12 & 156 (L) \\
         ChatGPT o3& 12 & 12 & 12 & 12 & 12 & 12 & 12 & 12 & 12 & 12 & 12 & 12 & 12 & 156 (L) \\
        \bottomrule
    \end{tabular}
    \label{tab:comparison}
\end{table*}

\section{Discussion and conclusion}

We tested the mathematical capabilities of different versions of LLMs using the Finnish secondary school matriculation examination in mathematics as test data. The study results demonstrated the rapid improvement of LLMs in performing upper secondary mathematics. The current capabilities of the best LLMs to elaborate and solve high-stakes mathematical problems are sufficient to achieve top performance in the Finnish national examinations.

The recent AI models employ various techniques for problem-solving, which significantly enhance their mathematical reasoning and problem-solving capabilities.  OpenAI’s \textit{o1} model \citep{o1systemcard2024} introduces a chain-of-thought (CoT) reasoning approach \citep{wei2022chain}, which mirrors the way a human might pause and reason before answering a complex question. Through reinforcement learning \citep{sutton2018reinforcement}, the model learns to structure its internal reasoning in sequential steps, improving its ability to identify and correct mistakes, simplify complex subproblems, and switch strategies when initial attempts fail \citep{openai2024reasoning}. OpenAI's o3-mini \citep{openai2025o3mini} introduced adjustable reasoning effort, i.e., it can adapt its depth of analysis using three reasoning effort settings (low, medium, and high). This flexibility enhances accuracy in mathematical problems, as seen in the AIME 2024 competition, where o3-mini achieved 87.3\% accuracy, outperforming GPT-4 (see Table \ref{tab:math_benchmark}) \citep{openai2025}.  

Another advancement in mathematical AI reasoning came from DeepSeek R1 \citep{deepseek2025huggingface}, which showed excellent performance and leverages reinforcement learning alone without human feedback-based fine-tuning to enhance multi-step logical reasoning \citep{guo2025}. While models such as GPT-4 rely on direct inference without iterative self-evaluation, DeepSeek R1 employs deliberative reasoning, where it revises and refines its reasoning process before finalizing an answer \citep{guo2025}. While o3-mini adjusts reasoning effort through predefined modes \citep{openai2025o3mini}, DeepSeek R1 learns to allocate reasoning steps adaptively through reinforcement learning~\citep{guo2025}. The combination of \textit{reinforcement learning, adaptive problem-solving, and self-correcting iteration} establishes DeepSeek R1 currently as one of the most advanced models in mathematical reasoning \citep{guo2025,mercer2025}.

Earlier studies remained quite skeptical about the mathematical capabilities of LLMs \citep{dao2023,frieder2023mathematical}. However, the recent development has significantly improved their usefulness in mathematics at the secondary education level and maybe even in the fields that utilize mathematics at higher education and in working life. In Finland, mathematics has the highest weight in certificate-based selection to higher education, influencing admissions to competitive fields such as medicine, law, and engineering \citep{hakanen2023}. According to the national higher education database \citep{vipunen}, the average grades in advanced mathematics for admitted students in 2023 were as follows: medicine 6.5, law 5.9, mathematics 5.6, and civil engineering 5.4.  For mathematics students, the grade distribution among those selected was: L (18\%), E (42\%), M (24\%), and C (16\%). Based on these statistics, the earlier versions of the LLMs as tested in Section \ref{subsec:GenAIPerf} already demonstrated a strong likelihood of qualifying for university studies in various fields, including mathematics itself.

Our results showed that nowadays, multiple LLMs achieve full or nearly full scores in the Finnish matriculation examination in advanced mathematics. This means that an up-to-date LLM would qualify for higher education and would likely be ranked among the top applicants in the most competitive fields. Similar results have been found in other studies. In the Turkish Dental Specialization Exam (DUS), OpenAI's o1 achieved 97.88\% accuracy (outperforming all human candidates), and Gemini 2.0 Advanced got 96.82\% \citep{kinikoglu2025}. Furthermore,  AI models have passed major medical licensing exams worldwide, including the USMLE (USA), PLAB (UK), and NMLE (Japan), which are not just entrance exams but full professional certification tests that determine the eligibility to practice medicine or start a specialization \citep{liu2024}.

While the present study demonstrates that state-of-the-art LLMs can now achieve top-level performance in high-stakes mathematics examinations, their potential as instructional agents remains an open question. As mathematical task-solving competence becomes increasingly saturated, future research should shift focus toward evaluating the pedagogical capabilities of LLMs in dialogue-based learning contexts. Specifically, it is essential to examine whether these models can engage in meaningful educational interactions — identifying student misconceptions, posing effective scaffolding questions, and fostering conceptual understanding without prematurely revealing solutions \citep{nino2024itsreview,tack2022aiteachertest}. Emerging datasets such as MATHDIAL \citep{Macina2023MathDial} offer promising frameworks for studying these pedagogical dimensions. Exploring the extent to which LLMs can function not only as problem solvers but also as adaptive and equitable tutors represents a critical direction for advancing their role in education.

\printbibliography

@String{Computing = "Computing" }

@article{sikstrom2024pedagogical,
  title={Pedagogical agents communicating and scaffolding students' learning: High school teachers' and students' perspectives},
  author={Sikstr{\"o}m, Pieta and Valentini, Chiara and Sivunen, Anu and K{\"a}rkk{\"a}inen, Tommi},
  journal={Computers \& Education},
  volume={222},
  pages={105140},
  year={2024},
  publisher={Elsevier},
  doi={10.1016/j.compedu.2024.105140}
}

@String{Springer = "Springer-Verlag" }

@article{ahn2024large,
  title={Large language models for mathematical reasoning: Progresses and challenges},
  author={Ahn, Janice and Verma, Rishu and Lou, Renze and Liu, Di and Zhang, Rui and Yin, Wenpeng},
  journal={arXiv preprint arXiv:2402.00157},
  year={2024},
  doi={https://doi.org/10.48550/arXiv.2402.00157}
}

@INPROCEEDINGS{Heilala2025-wg,
  title     = "Beyond Text-to-Text: An Overview of Multimodal and Generative
               Artificial Intelligence for Education Using Topic Modeling",
  author    = "Heilala, Ville and Araya, Roberto and Hämäläinen, Raija",
  booktitle = "Proceedings of the 40th ACM/SIGAPP Symposium on Applied Computing
               (SAC '25)",
  publisher = "ACM",
  address   = "New York, NY, USA",
  pages     = "1--10",
  year      =  2025,
  doi       = "10.1145/3672608.3707764"
}

@misc{YoKoeGrading,
  title        = "Characteristics of good answers",
  author       = "{The Matriculation Examination Board of Finland}",
  url = "https://www.ylioppilastutkinto.fi/fi/tutkinnon-suorittaminen/hyvan-vastauksen-piirteet",
  year         = 2025,
  note         = "Accessed: February 21, 2025 (in Finnish)"
}

@article{mercer2025,
  author    = {Mercer, S. and Spillard, S. and Martin, D. P.},
  title     = {Brief Analysis of DeepSeek R1 and Its Implications for Generative AI},
  journal   = {arXiv preprint arXiv:2502.02523},
  year      = {2025},
  url = {https://doi.org/10.48550/arXiv.2502.02523}
}

@article{guo2025,
  author    = {Guo, D. and Yang, D. and Zhang, H. and Song, J. and Zhang, R. and Xu, R. and He, Y.},
  title     = {DeepSeek-R1: Incentivizing Reasoning Capability in LLMs via Reinforcement Learning},
  journal   = {arXiv preprint arXiv:2501.12948},
  year      = {2025},
  url = {https://doi.org/10.48550/arXiv.2501.12948}
}

@misc{openai2025,
  author    = {OpenAI},
  title     = {OpenAI o3-mini: Pushing the frontier of cost-effective reasoning},
  howpublished = {\url{https://openai.com/index/openai-o3-mini/}},
  year      = {2025},
  note      = {Accessed: February 9, 2025}
}

@article{liu2024,
  author    = {Liu, M. and Okuhara, T. and Chang, X. and Shirabe, R. and Nishiie, Y. and Okada, H. and Kiuchi, T.},
  title     = {Performance of ChatGPT across different versions in medical licensing examinations worldwide: systematic review and meta-analysis},
  journal   = {Journal of Medical Internet Research},
  volume    = {26},
  pages     = {e60807},
  year      = {2024},
  doi       = {10.2196/60807}
}

@article{kinikoglu2025,
  author    = {Kinikoglu, I.},
  title     = {Evaluating ChatGPT and Google Gemini performance and implications in Turkish dental education},
  journal   = {Cureus},
  volume    = {17},
  number    = {1},
  year      = {2025},
  doi       = {10.7759/cureus.77292}
}

@article{almarashdi2024unveiling,
  author    = {H. S. Almarashdi and A. M. Jarrah and O. A. Khurma and S. M. Gningue},
  title     = {Unveiling the potential: A systematic review of ChatGPT in transforming mathematics teaching and learning},
  journal   = {EURASIA Journal of Mathematics, Science and Technology Education},
  volume    = {20},
  number    = {12},
  pages     = {em2555},
  year      = {2024},
  doi={10.29333/ejmste/15739}
}

@misc{maa_invitationals,
  author    = {{Mathematical Association of America (MAA)}},
  title     = {MAA Invitational Competitions},
  year      = {2024},
  url       = {https://maa.org/maa-invitational-competitions/},
  note      = {Accessed: 2024-02-05}
}

@article{mirzadeh2024gsm,
  author    = {I. Mirzadeh and K. Alizadeh and H. Shahrokhi and O. Tuzel and S. Bengio and M. Farajtabar},
  title     = {Gsm-symbolic: Understanding the limitations of mathematical reasoning in large language models},
  journal   = {arXiv preprint arXiv:2410.05229},
  year      = {2024},
  url       = {https://doi.org/10.48550/arXiv.2410.05229}
}

@article{kaplan2025chatgpt,
  author    = {H. A. Kaplan},
  title     = {ChatGPT's Knowledge in Mathematics Teaching: An Example of Rational Numbers},
  journal   = {Pegem Journal of Education and Instruction},
  volume    = {15},
  number    = {2},
  pages     = {63-75},
  year      = {2025},
  doi = {10.47750/pegegog.15.02.07}
}

@misc{openai2025o3mini,
  author       = {OpenAI},
  title        = {OpenAI o3-mini: Pushing the frontier of cost-effective reasoning},
  year         = {2025},
  howpublished = {\url{https://openai.com/index/openai-o3-mini/}},
  note         = {Accessed: 11 April 2025}
}

@misc{openai2024reasoning,
  author       = {OpenAI},
  title        = {Learning to Reason with Language Models},
  year         = {2024},
  howpublished = {\url{https://openai.com/index/learning-to-reason-with-llms/}},
  note         = {Accessed: 11 April 2025}
}

@article{hendrycks2021math,
  author    = {D. Hendrycks and C. Burns and S. Kadavath and A. Arora and S. Basart and E. Tang and J. Steinhardt},
  title     = {Measuring mathematical problem solving with the math dataset},
  journal   = {arXiv preprint arXiv:2103.03874},
  year      = {2021},
  url = {https://doi.org/10.48550/arXiv.2103.03874}
}

@article{achiam2023gpt,
  title={{GPT}-4 technical report},
  author={Achiam, Josh and Adler, Steven and Agarwal, Sandhini and Ahmad, Lama and Akkaya, Ilge and Aleman, Florencia Leoni and Almeida, Diogo and Altenschmidt, Janko and Altman, Sam and Anadkat, Shyamal and others},
  journal={arXiv preprint arXiv:2303.08774},
  year={2023},
  doi={https://doi.org/10.48550/arXiv.2303.08774}
}

@article{sanchez2023chatgpt,
  title={{ChatGPT} challenges blended learning methodologies in engineering education: A case study in mathematics},
  author={S{\'a}nchez-Ruiz, Luis M and Moll-L{\'o}pez, Santiago and Nu{\~n}ez-P{\'e}rez, Adolfo and Mora{\~n}o-Fern{\'a}ndez, Jos{\'e} Antonio and Vega-Fleitas, Erika},
  journal={Applied Sciences},
  volume={13},
  number={10},
  pages={6039},
  year={2023},
  publisher={MDPI},
  doi={10.3390/app13106039}
}

@article{wardat2023chatgpt,
  title={{ChatGPT}: A revolutionary tool for teaching and learning mathematics},
  author={Wardat, Yousef and Tashtoush, Mohammad A and AlAli, Rommel and Jarrah, Adeeb M},
  journal={Eurasia Journal of Mathematics, Science and Technology Education},
  volume={19},
  number={7},
  pages={em2286},
  year={2023},
  publisher={Modestum},
  doi={10.29333/ejmste/13272}
}

@article{zhao2023survey,
  title={A survey of large language models},
  author={Zhao, Wayne Xin and Zhou, Kun and Li, Junyi and Tang, Tianyi and Wang, Xiaolei and Hou, Yupeng and Min, Yingqian and Zhang, Beichen and Zhang, Junjie and Dong, Zican and others},
  journal={arXiv preprint arXiv:2303.18223},
  year={2023},
  url={https://doi.org/10.48550/arXiv.2303.18223}
}

@article{frieder2023mathematical,
  title={Mathematical capabilities of {ChatGPT}},
  author={Frieder, Simon and Pinchetti, Luca and Chevalier, Alexis and Griffiths, Ryan-Rhys and Salvatori, Tommaso and Lukasiewicz, Thomas and Petersen, Philipp Christian and Berner, Julius},
  journal={arXiv preprint arXiv:2301.13867},
  year={2023},
  url={https://doi.org/10.48550/arXiv.2301.13867}
}

@article{urhan2024argumentation,
  title={An argumentation experience regarding concepts of calculus with {ChatGPT}},
  author={Urhan, Selin and Gen{\c{c}}aslan, O{\u{g}}uzhan and Dost, {\c{S}}enol},
  journal={Interactive Learning Environments},
  pages={1--26},
  year={2024},
  publisher={Taylor \& Francis},
  doi={10.1080/10494820.2024.2308093}
}

@ARTICLE{kasneci2023chatgpt,
  title   = "{ChatGPT} for good? On opportunities and challenges of large
             language models for education",
  author  = "Kasneci, Enkelejda and Sessler, Kathrin and Küchemann, Stefan and
             Bannert, Maria and Dementieva, Daryna and Fischer, Frank and
             Gasser, Urs and Groh, Georg and Günnemann, Stephan and Hüllermeier,
             Eyke and Krusche, Stephan and Kutyniok, Gitta and Michaeli, Tilman
             and Nerdel, Claudia and Pfeffer, Jürgen and Poquet, Oleksandra and
             Sailer, Michael and Schmidt, Albrecht and Seidel, Tina and Stadler,
             Matthias and Weller, Jochen and Kuhn, Jochen and Kasneci, Gjergji",
  journal = "Learning and individual differences",
  volume  =  103,
  pages   =  102274,
  month   =  apr,
  year    =  2023,
  doi     = "10.1016/j.lindif.2023.102274",
  issn    = "1041-6080"
}

@inproceedings{dasari2024chatgpt,
  title={{ChatGPT} in didactical tetrahedron, does it make an exception? A case study in mathematics teaching and learning},
  author={Dasari, Dadan and Hendriyanto, Agus and Sahara, Sani and Suryadi, Didi and Muhaimin, Lukman Hakim and Chao, Theodore and Fitriana, Laila},
  booktitle={Frontiers in Education},
  volume={8},
  pages={1295413},
  year={2024},
  organization={Frontiers},
  doi={10.3389/feduc.2023.1295413}
}

@article{siegle2023twenty,
  title={Twenty-five years of learning with pedagogical agents: History, barriers, and opportunities},
  author={Siegle, Robert F and Schroeder, Noah L and Lane, H Chad and Craig, Scotty D},
  journal={TechTrends},
  volume={67},
  number={5},
  pages={851--864},
  year={2023},
  publisher={Springer},
  doi={10.1007/s11528-023-00869-3}
}

@article{sikstrom2022pedagogical,
  title={How pedagogical agents communicate with students: A two-phase systematic review},
  author={Sikstr{\"o}m, Pieta and Valentini, Chiara and Sivunen, Anu and K{\"a}rkk{\"a}inen, Tommi},
  journal={Computers \& Education},
  volume={188},
  pages={104564},
  year={2022},
  publisher={Elsevier},
  doi={10.1016/j.compedu.2022.104564}
}

@article{naveed2023comprehensive,
  title={A comprehensive overview of large language models},
  author={Naveed, Humza and Khan, Asad Ullah and Qiu, Shi and Saqib, Muhammad and Anwar, Saeed and Usman, Muhammad and Barnes, Nick and Mian, Ajmal},
  journal={arXiv preprint arXiv:2307.06435 (version v9 from Apr, 2024)},
  year={2023},
  url={https://doi.org/10.48550/arXiv.2307.06435}
}

@book{OECDDigitalization2023,
   author = "OECD",
   title = "{OECD} Digital Education Outlook 2023",
   year = "2023",
   pages = 410,
   url = "https://www.oecd-ilibrary.org/content/publication/c74f03de-en",
   doi = "https://doi.org/https://doi.org/10.1787/c74f03de-en" 
}

@article{virtanen2024vector,
  title={Vector Misconceptions in {F}innish Matriculation Examination},
  author={Virtanen, Heli and Ernvall-Hyt{\"o}nen, Anne-Maria and Laaksonen, Antti},
  journal={FMSERA Journal},
  pages={1--17},
  year={2024},
  url={https://journal.fi/fmsera/article/view/127798}
}

@article{pesovski2024generative,
  title={Generative {AI} for Customizable Learning Experiences},
  author={Pesovski, Ivica and Santos, Ricardo and Henriques, Roberto and Trajkovik, Vladimir},
  journal={Sustainability},
  volume={16},
  number={7},
  pages={3034},
  year={2024},
  publisher={MDPI},
  doi={10.3390/su16073034}
}

@article{dao2023,
  title={Investigating the effectiveness of {chatGPT} in mathematical reasoning and problem solving: Evidence from the {V}ietnamese national high school graduation examination},
  author={Dao, X. Q. and Le, N. B.},
  journal={arXiv preprint arXiv:2306.06331},
  year={2023},
  url={https://doi.org/10.48550/arXiv.2306.06331}
}

@techreport{hakanen2023,
  title={Todistusvalintahankkeen loppuraportti},
  author={Hakanen, M. and Kalmbach, A. and Kuuppelomäki, T. and Lauronen, T. and Suhonen, T. and Virkola, T.},
  year={2023},
  institution = {{VATT} Muistiot 70, {VATT} Institute for Economic Research},
  note = {(in Finnish)},
  url = {https://urn.fi/URN:ISBN:978-952-274-293-3}
}

@misc{vipunen,
  author       = {{Vipunen}},
  title        = {Vipunen - Education Statistics Finland},
  year         = {2024},
  howpublished = {\url{https://vipunen.fi/en-gb/}},
  note         = {Accessed: 2024-06-27}
}

@article{Macina2023MathDial,
  author       = {Macina, Johannes and Daheim, Nico and Chowdhury, Sreyash P. and Sinha, Tanmay and Kapur, Manu and Gurevych, Iryna and Sachan, Mrinmaya},
  title        = {{MathDial: A Dialogue Tutoring Dataset with Rich Pedagogical Properties Grounded in Math Reasoning Problems}},
  journal      = {arXiv preprint arXiv:2305.14536},
  year         = {2023},
  url          = {https://doi.org/10.48550/arXiv.2305.14536}
}

@article{tack2022aiteachertest,
  title={The AI Teacher Test: Measuring the Pedagogical Ability of Blender and GPT-3 in Educational Dialogues},
  author={Tack, Anaïs and Piech, Chris},
  journal={arXiv preprint arXiv:2205.07540},
  year={2022},
  url={https://doi.org/10.48550/arXiv.2205.07540}
}

@article{nino2024itsreview,
  author = {Niño-Rojas, F. and Lancheros-Cuesta, D. and Jiménez-Valderrama, M. T. P. and Mestre, G. and Gómez, S.},
  title = {Systematic Review: Trends in Intelligent Tutoring Systems in Mathematics Teaching and Learning},
  journal = {International Journal of Education in Mathematics, Science and Technology},
  year = {2024},
  volume = {12},
  number = {1},
  pages = {203--229},
  doi = {10.46328/ijemst.3189},
  url = {https://doi.org/10.46328/ijemst.3189}
}

@misc{o1systemcard2024,
  author       = {OpenAI},
  title        = {OpenAI o1 System Card},
  year         = {2024},
  howpublished = {\url{https://cdn.openai.com/o1-system-card-20241205.pdf}},
  note         = {Accessed: 11 April 2025}
}

@article{wei2022chain,
  title={Chain-of-thought prompting elicits reasoning in large language models},
  author={Jason Wei and Xuezhi Wang and Dale Schuurmans and Maarten Bosma and Ed Chi and Quoc Le and Denny Zhou},
  journal={arXiv preprint arXiv:2201.11903},
  year={2022},
  url={https://doi.org/10.48550/arXiv.2201.11903}
}

@book{sutton2018reinforcement,
  title={Reinforcement Learning: An Introduction},
  author={Sutton, Richard S and Barto, Andrew G},
  year={2018},
  publisher={MIT Press},
  edition={2nd}
}

@misc{deepseek2025huggingface,
  author       = "{DeepSeek AI}",
  title        = {DeepSeek R1 Release – Hugging Face Model Card},
  year         = {2025},
  howpublished = {\url{https://huggingface.co/deepseek-ai/DeepSeek-R1}},
  note         = {Accessed: 11 April 2025}
}

@article{corbin2025missing,
  title={The missing story of GenAI summarisers: a critical research agenda},
  author={Corbin, Thomas Alexander and Walton, Jack},
  journal={Higher Education Research \& Development},
  pages={1--14},
  year={2025},
  publisher={Taylor \& Francis}
}

\end{document}